\pdfoutput=1

\documentclass[11pt]{article}

\usepackage[]{acl}

\usepackage{times}
\usepackage{latexsym}
\usepackage{graphicx}
\usepackage{amsmath}
\usepackage[T1]{fontenc}
\usepackage{arydshln}
\usepackage[utf8]{inputenc}
\usepackage{subfigure}
\usepackage{amssymb}
\usepackage{paralist}
\usepackage{arydshln}

\usepackage{microtype}
\usepackage{multirow}
\usepackage{color}

\newcommand{\CLS}{\mbox{CLS}}
\usepackage{algorithm}
\usepackage{algorithmic}

\usepackage{amsmath,amsfonts,bm}

\def\eqref#1{equation~\ref{#1}}

\def\1{\bm{1}}

\DeclareMathAlphabet{\mathsfit}{\encodingdefault}{\sfdefault}{m}{sl}
\SetMathAlphabet{\mathsfit}{bold}{\encodingdefault}{\sfdefault}{bx}{n}

\title{Trigger-free Event Detection via Derangement Reading Comprehension}

\author{Jiachen Zhao \\
  HKUST \\
  Hong Kong, China \\
  \texttt{jzhaobc@connect.ust.hk} \\\And
  
  Haiqin Yang \\
  International Digital Economy Academy \\
  Shenzhen, China\\
  \texttt{hqyang@ieee.org} \\}

\begin{document}
\maketitle 
\begin{abstract} 
Event detection (ED), aiming to detect events from texts and categorize them, is vital to understanding actual happenings in real life.  However, mainstream event detection models require high-quality expert human annotations of triggers, which are often costly and thus deter the application of ED to new domains.  Therefore, in this paper, we focus on low-resource ED without triggers and aim to tackle the following formidable challenges: multi-label classification, insufficient clues, and imbalanced events distribution. We propose a novel trigger-free ED method via Derangement mechanism on a machine Reading Comprehension (DRC) framework.  More specifically, we treat the input text as {\em Context} and concatenate it with all event type tokens that are deemed as {\em Answers} with an omitted default question.  So we can leverage the self-attention in pre-trained language models to absorb semantic relations between input text and the event types.  Moreover, we design a simple yet effective event derangement module (EDM) to prevent major events from being excessively learned so as to yield a more balanced training process.  The experiment results show that our proposed trigger-free ED model is remarkably competitive to mainstream trigger-based models, showing its strong performance on low-source event detection.  
\end{abstract} 

\section{Introduction} 
\label{sec:intro}
The task of event detection (ED), aiming to spot the appearance of predefined event types from texts and classify them, is vital to {understanding the actual happenings in real life~\cite{DBLP:phd/hal/Edouard17,DBLP:journals/grid/SaeedAMSRDAX19}.}  Taking an example from the Automatic Context Extraction (ACE) corpus:
\begin{quote}
{\bf S:} And they sent him to Baghdad and killed. 
\label{quote:example}
\end{quote}
This sentence consists of two events, {\em Transport} and {\em Die}.  A desired ED system should correctly identify these two events simultaneously.  At first glance, this task can be arduous and challenging because event types implicitly exist in sentences. Therefore, in the literature, researchers usually tackle this problem via a two-stage trigger-based framework.  Triggers, i.e., words or phrases providing the \emph{most clear} indication of an event occurrence, are first identified and then events are recognized  accordingly~\cite{ahn2006stages,DBLP:conf/acl/LiJH13,DBLP:conf/acl/ChenXLZ015}.  For example, in the above example, ``sent'' and ``killed'' are the corresponding triggers for the event of {\em Transport} and {\em Die}, respectively.  

{However, trigger-based models are difficult to transfer to new domains because they require massive high-quality triggers annotation~\cite{DBLP:conf/acl/LaiND20,lu-etal-2021-text2event}.  The annotation process is usually expensive and time-consuming~\cite{shen-etal-2021-corpus}, which may take linguistics experts multiple rounds to screen the data~\cite{doddington-etal-2004-automatic}.  An effective but challenging solution to arduous labeling is event detection without triggers~\cite{zeng2018scale,DBLP:conf/naacl/LiuLZYZ19,zheng-etal-2019-doc2edag}.  However, such a forbidding \textit{low-resource} setting leads to insufficient clues, making trigger-free approaches inferior to trigger-based methods~\cite{DBLP:conf/naacl/LiuLZYZ19}.  As we hope to extend ED to more domains with less human effort in real-world applications, a key issue then rises, \textit{how to design a trigger-free ED model that is competitive with trigger-based models?}}

To fill the gaps, we focus on event detection without triggers to reduce strenuous effort in data labeling.  We aim at tackling the following formidable challenges: (1) {\bf Multi-label issue:} each input sentence may hold zero or multiple events, which can be formulated into a challenging multi-label classification task. (2) {\bf Insufficient clues:} triggers are of significance to detect events~\cite{DBLP:conf/acl/ZhangKLMH20,DBLP:conf/acl/EbnerXCRD20}.  Without explicitly annotated triggers, we may lack sufficient clues to identify the event types and need to {seek alternatives to shed light on the correlation between words and the event types.} (3) {\bf Imbalanced events distribution:} as shown in Fig.~\ref{fig:data}, events may follow the Matthew effect.  Some events dominate the data while others contain only several instances.  The imbalanced event distribution brings significant obstacles to learn and detect minor events.  

\begin{figure*}[htp]
\centering
\includegraphics[scale=0.47]{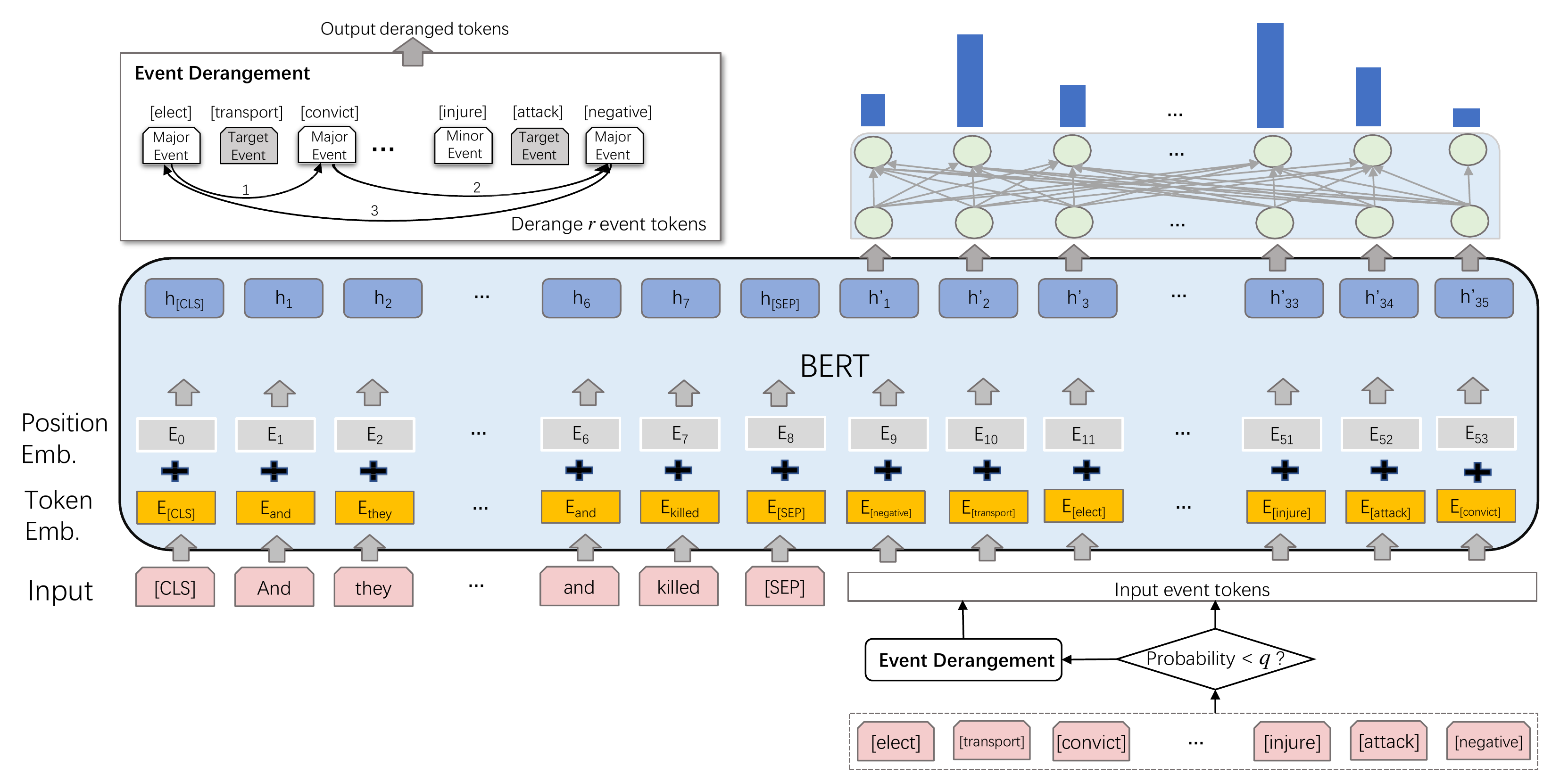}
\caption{Our proposed DRC is on top of BERT.  It consists of three main modules: RC encoder, the event derangement module (EDM), and the multi-label classifier.  The EDM is amplified in the upper-left corner for better illustration; see more description in the main text. 
}
\label{fig:model2}
\end{figure*}

Hence, we propose a trigger-free ED method via Derangement mechanism on a machine Reading Comprehension (DRC) framework to tackle the challenges.  Figure~\ref{fig:model2} illustrates our proposed framework with three main modules: the RC encoder, the event derangement module (EDM), and the multi-label classifier.  {In the RC encoder, the input sentence, deemed as ``Context'', and all event tokens, appended as ``Answers'', are fed into BERT~\cite{DBLP:conf/naacl/DevlinCLT19} simultaneously.  Such design allows the model to absorb all information without explicitly indicating triggers and enables the model to automatically learn helpful semantic relations between input texts and event tokens through the self-attention mechanism of Transformer~\cite{DBLP:conf/nips/VaswaniSPUJGKP17}.  During training, the EDM is activated with a certain probability only when the ground-truth events are major events.  By perturbing the order of other event tokens, the model can prevent excessive update on the major events, which implicitly under-samples the training instances in the major events and thus yields a more balanced training process.}  Finally, the learned contextual representations of event tokens are fed into a multi-label classifier to produce the probabilities of each event type in the input text.  

In summary, the contribution of our work is threefold: (1) we propose a competitive paradigm to an important task, namely multi-label event detection without triggers.  Through a simplified machine reading comprehension framework, we can directly capture the semantic relation between input texts and event types without explicitly annotated triggers.  (2) During training, we implement a simple yet effective module, i.e., the event derangement module, to overcome the imbalanced learning issue.  By perturbing the order of event tokens, we prevent major events from being excessively learned by model, which can make the training process more balanced.
(3)  We report that our proposal can achieve the state-of-the-art performance on event detection on public benchmark datasets.  Further gradient explanation also indicates that our trigger-free model can spot and link triggers to the corresponding events by itself and identify related event arguments as well.

\section{Related Work} 
Existing event detection methods usually rely on triggers to detect events.  
These approaches need to explicitly identify triggers and assign them with predefined event types afterwards.  For example, in~\cite{DBLP:conf/acl/LiJH13}, structured Perceptron has been exploited on hand-made features to identify triggers. In~\cite{DBLP:conf/naacl/NguyenCG16}, triggers and arguments are jointly identified by utilizing bidirectional recurrent neural networks.
Additionally, some approaches formulate the task as machine reading comprehension or question answering~\cite{DBLP:conf/emnlp/DuC20,DBLP:conf/emnlp/LiuCLBL20}. A predefined question template concatenating with the input sentence will be fed into a language model to identify the corresponding triggers. Searching optimal results from multiple predefined templates may be needed during inference.  

To reduce massive annotation of triggers in new domains, more and more proposals have reformulated event detection as a {\em low-resource} natural language processing  task by discarding some information~\cite{DBLP:conf/naacl/LiuLZYZ19,lu-etal-2021-text2event} or only using partial data~\cite{DBLP:conf/acl/LaiND20,hsu2022degree}.  Event detection without triggers is proposed in~\cite{DBLP:conf/naacl/LiuLZYZ19} which incorporates features of event types to input sentences via the attention mechanism on an LSTM model.  The proposal does not utilize recently-developed pre-trained language models and yields much worse performance than trigger-based methods.  Moreover, it splits a multi-label event detection task into multiple binary classification tasks for each event, which leads to slow inference.  
Most recently, \citet{lu-etal-2021-text2event} proposed a sequence-to-structure framework to extract events in an end-to-end manner.  Though it makes token-level annotations (i.e., trigger offset) unnecessary, the proposed method still requires strenuous effort to specify trigger words.  On the other hand, some approaches can attain satisfactory results with partial raw data labeled~\cite{DBLP:conf/acl/LaiND20,hsu2022degree}.  For example,~\citet{DBLP:conf/acl/LaiND20} adopt a prototypical framework to classify representative vectors computed from the embeddings of input texts.  \citet{hsu2022degree} feed input texts and a manually designed prompt into a generation-based neural network to output a natural sentence for further event extraction.  Nevertheless, these approaches are less useful with more raw data accessible since parts of the data may be left unemployed for reducing annotation.

\section{Methodology}
\subsection{Task Definition}
\label{sec:def}
Following~\cite{ahn2006stages,DBLP:conf/acl/JiG08,DBLP:conf/naacl/LiuLZYZ19}, we are given a set of training data, $\{(x_i, y_i)\}_{i=1}^N$, where $N$ is the number of sentence-event pairs.  $x_i=w_{i1}w_{i2}\ldots w_{i|x_i|}$ is the $i$-th sentence with $|x_i|$ tokens and $y_i\subseteq\mathcal{S}$ is an event set, which records all related event(s).  $\mathcal{S}=\{e_1, e_2, \ldots, e_n\}$ consists of all $n$ events, including an additional ``negative'' event meaning that sentences do not contain any events.  Our goal is to train a model to detect the corresponding event type(s) as accurate as possible given an input sentence.  This can be formulated as the multi-label classification task in machine learning.  Our tasks lie in (1) how to learn more precise representations to embed the semantic information between texts and event types? (2) How to deliver the multi-task classification task effectively?

\paragraph{Major Events vs. Minor Events} Imbalanced event distribution is a major issue in our setting.   Traditionally, Imbalance Ratio (IR)~\cite{DBLP:journals/tsmc/GalarFTSH12} is a typical metric to estimate the imbalance of the data.  However, {IR provides little distribution information about the middle classes}~\cite{ortigosa2017measuring}.  To articulate the differences of major events and minor events, we borrow its definition in~\cite{dong2018imbalanced} to distinguish them.  We first sort all event types in descending order with respect to the number of instances in each class and obtain the sorted sequence:
\begin{equation}\label{eq:order_seq}
S_{\tiny\mbox{SA}}= e_1\ldots e_{n},\quad  \mbox{where~~} |{e_{i}}| \ge |{e_{i+1}}|.
\end{equation} 
Here, $e_{i}$ denotes the $i$-th event type with $|{e_{i}}|$ instances.  

Then, we define the set of major events as the top-$k$ elements in $S_{\tiny\mbox{SA}}$ while the remaining elements as the minor events:
\begin{eqnarray}
&&E_{\tiny\mbox{Major}}=\{ e_{i}\;|\;i=1,2,...k\}, \\
&& E_{\tiny\mbox{Minor}}=\{ e_{i}\;|\;i=k+1,...n\},
\end{eqnarray}
{where $k$ is determined by a hyperparamter of $\alpha$ by rounding to the nearest integer if it is a float.  Here, $\alpha$ indicates the percentage of the major events in all $N$ sentence-event pairs:} 
\begin{equation}\\\label{eq:major_class_definition}
\alpha*N=\sum_{i=1}^{k} |e_{i}|.  
\end{equation}
{Usually, $\alpha$ is simply set to 0.5 as~\cite{dong2018imbalanced}. }

\subsection{Our Proposal}
\label{sec:method}
{Figure~\ref{fig:model2} outlines the overall structure of our proposed DRC, which consists of three main modules: the RC encoder, the multi-label classifier, and the event derangement module (EDM).}

\begin{algorithm}[htp]  
\caption{Event Derangement}
\label{alg:EDM}   
\begin{algorithmic}[1] 
\REQUIRE  Input sentence $x$; The initial event sequence $S_{\tiny{\mbox{init}}}$; The sequence of all event types in descending order $S_{\tiny\mbox{SA}}$; Possibility $q$; Number $r$\\
\ENSURE Deranged sequence of event tokens $S_{\tiny\mbox{O}}$
\STATE Initialize $E_{\tiny\mbox{GT}}$ as the set of the ground truth event types implied by $x$
\STATE Initialize $E_{\tiny\mbox{D}}$ with $r$ events that are not in $E_{\tiny\mbox{GT}}$ from the beginning of $S_{\tiny\mbox{SA}}$  
\STATE Initialize $E_{tmp}=\emptyset$ \texttt{\small\em \# a helper set~to record the selected event types during derangement}
\STATE Initialize $S_{\tiny\mbox{O}}=[]$
\STATE Generate $rand$ uniformly from [0, 1] 
\IF{$E_{\tiny\mbox{GT}}\cap E_{\tiny\mbox{Major}}\neq \emptyset$ and $rand$ < $q$}
\FOR {$e_{curr}$ in $S_{\tiny{\mbox{init}}}$}
\IF {$e_{curr}$ in $E_{\tiny\mbox{D}}$}
\STATE Randomly select $e$ from $E_{\tiny\mbox{D}}$ and $e\neq e_{curr}$ and $e\notin E_{tmp}$
\STATE Append $e$ to $S_{\tiny\mbox{O}}$
\STATE Add $e$ to $E_{tmp}$
\ELSE \STATE{Append $e_{curr}$ to $S_{\tiny\mbox{O}}$}
\ENDIF
\ENDFOR
\ELSE \STATE $S_{\tiny\mbox{O}}=S_{\tiny{\mbox{init}}}$
\ENDIF
\STATE Return $S_{\tiny\mbox{O}}$
\end{algorithmic}  
\end{algorithm}
\paragraph{RC Encoder} 
Our proposal is based on BERT due to its power in learning the contextual representation in the sequence of tokens~\cite{DBLP:conf/naacl/DevlinCLT19}.  {We present a simplified machine reading comprehension (MRC) framework: 
\begin{quote}
\textbf{[CLS] Context [SEP] Answers }
\end{quote}
where Context is the input sentence and Answers sequence all the event types.  This setup is close to MRC with the multiple choices option.  That is, it views the input sentence as Context and event types as the multiple choices (or Answers) with an omitted default question: ``What is the event type/what are the event types in the Context?''.  With both input texts and event tokens as the input of BERT, we can utilize BERT to automatically capture the relation between input texts and event types without explicitly indicating the triggers.}


In the implementation, given a training set, we first generate a {random} event order index $I_{\tiny\mbox{init}}=s_1\ldots s_n$, which is a permutation of $\{1, \ldots, n\}$, and obtain its initial sequence of event tokens  $S_{\tiny\mbox{init}}=e_{s_1}\ldots e_{s_n}$.  The event sequence $S_{\tiny\mbox{init}}$ is kept fixed for both training and testing.  Hence, given a sentence $x=w_1 \ldots w_{|x|}$, we obtain
\begin{equation}
\mbox{Input}\!=\![\mbox{CLS}]\, w_1\, \ldots \,w_{|x|}\,[\mbox{SEP}]\,e_{s_{1}}\, \ldots\,e_{s_{n}}. 
\end{equation}
{To avoid word-piece segmentation, we employ a square bracket around an event type, e.g., the event token of {\em Transport} is converted to ``[Transport]''.  This allows us to learn more precise event token representations and yield better performance (experiment results shown in Appendix~\ref{sec:wdembed}).  {Additionally, we apply position embeddings to event tokens based on their order in $S_{\tiny\mbox{init}}$ following the standard setup of BERT.  This can make BERT order-sensitive and further help BERT distinguish event types; see more discussion in Sec.~\ref{sec:exp_event_sequence} }}  

After that, we learn the hidden representations:
\begin{align}\nonumber
h_{\tiny{[\CLS]}}, h_{1}^{w}, \ldots, &~ h_{|x|}^{w}, h_{\tiny{[\mbox{SEP}]}}, h_{1}^{e}, \ldots, h_{n}^{e}\\=&~\text{BERT}(\mbox{Input}),
\end{align}
where $h_{i}^{w}$ is the hidden state of the $i$-{th} input token and $h_{i}^{e}$ is the hidden state of the corresponding event type, namely $e_{s_{i}}$.

\begin{figure*}[htp]
\centering
\subfigure[Event substypes on ACE2005]{\includegraphics[scale=0.33]{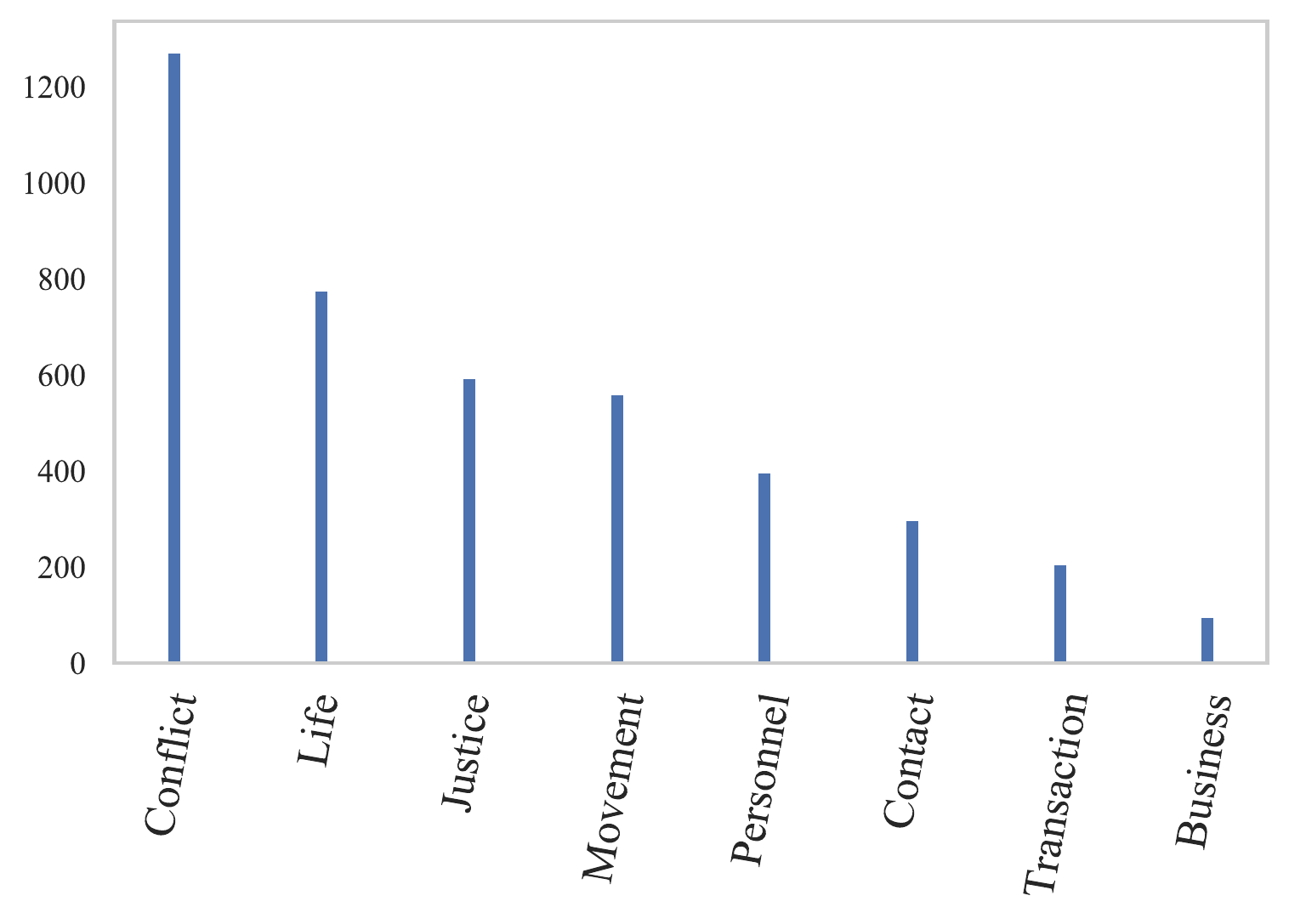}\label{fig:ace_second_level}}
\subfigure[Event main types on ACE2005]{\includegraphics[scale=0.33]{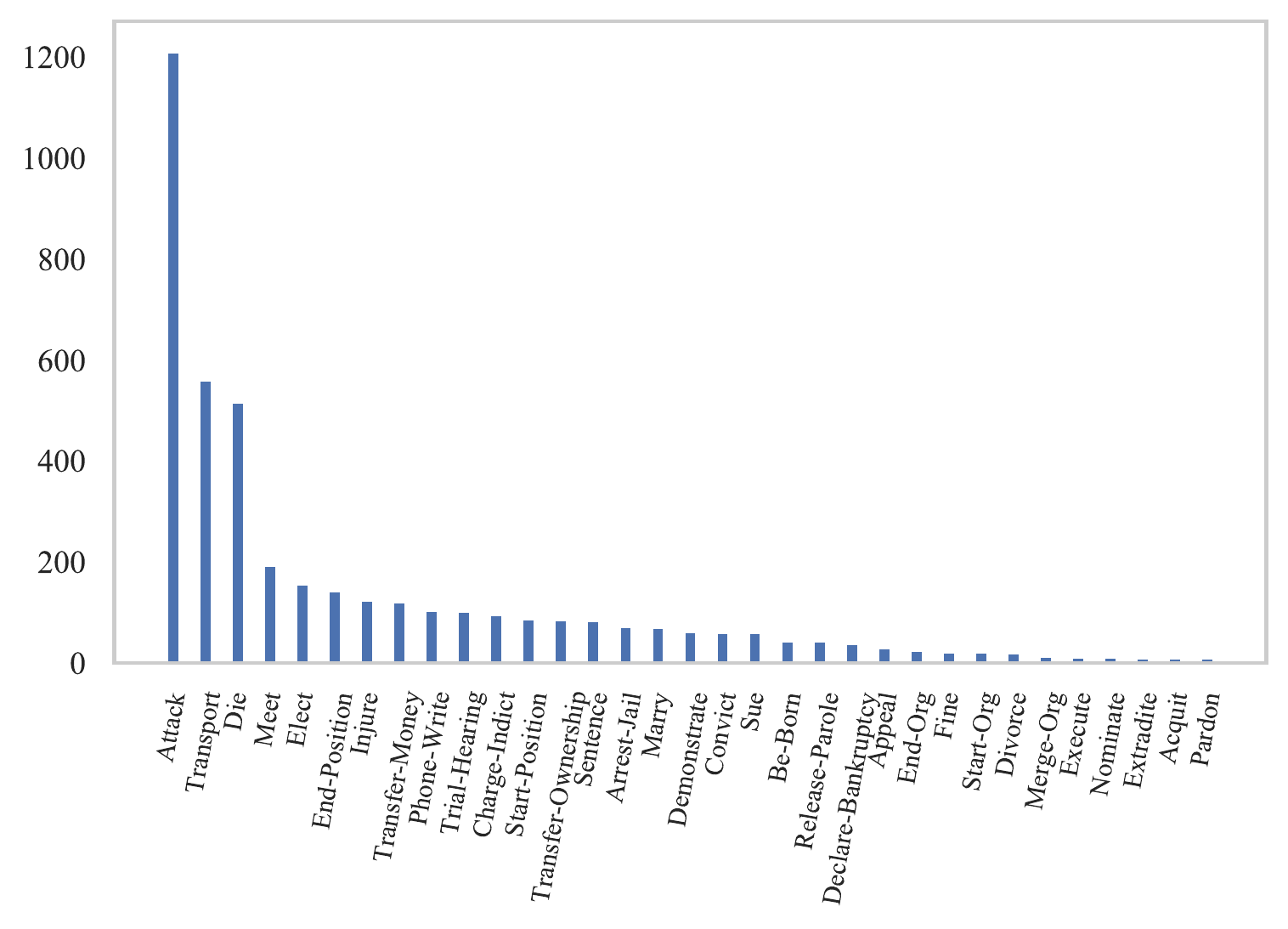}\label{fig:ace_first_level}}
\subfigure[Event types on TAC-KBP-2015]{\includegraphics[scale=0.33]{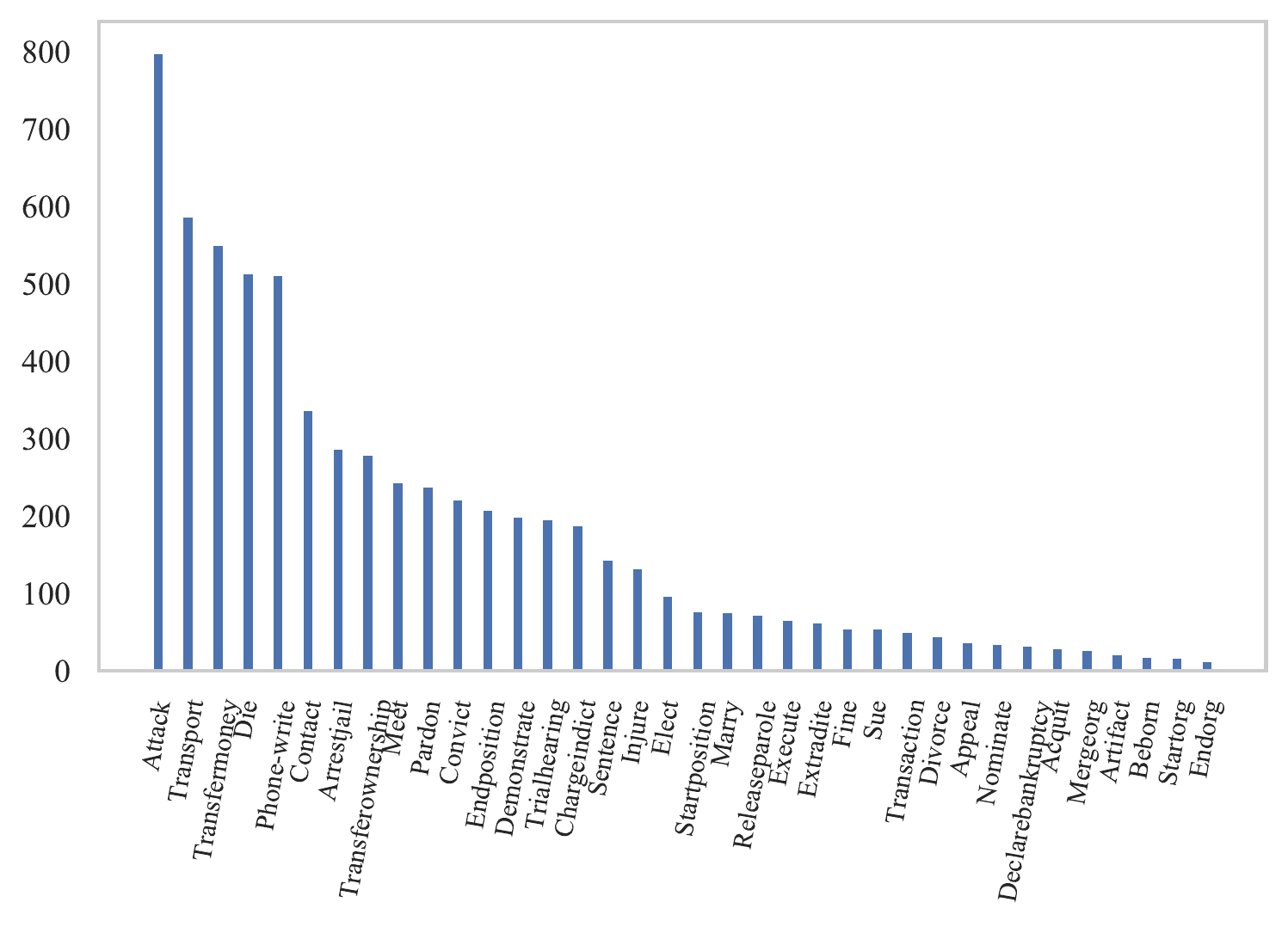}\label{fig:kbp_second_level}}
\caption{The event distributions of ACE2005 and TAC-KBP-2015 are intrinsically long-tailed.\label{fig:data}}
\end{figure*}

\paragraph{Multi-label Classifier} After learning the contextualized representations of the $\mbox{Input}$, we turn to construct the multi-label classifier.  Traditional methods usually apply a Multi-Layer Perception (MLP) on the [CLS] token to yield the classifier.  Differently, we feed all hidden states of event types to an MLP for the classification due to the supportive evaluation in Appendix~\ref{sec:App_Input_Classifier}.  Hence, given an input sentence $x$, we compute the predicted probability for the corresponding events by 
\begin{equation}
\hat{p}=\sigma\left(\text{MLP}\left ( h_{1}^{e},...,h_{n}^{e} \right)\right).
\end{equation}
{Since $\hat{p}$ is normalized to the range of 0 and 1, for simplicity, we follow~\cite{DBLP:conf/naacl/LiuLZYZ19} to determine the event labels when $\hat{p}\ge 0.5$.}

Our model can then be trained by minimizing the following loss:
\begin{equation} 
    \mathcal{L}\propto -\sum_{i=1}^N\sum_{j=1}^{n}(p_{ij}\log(\hat{p}_{ij})+(1-p_{ij})\log(1-\hat{p}_{ij}))
\end{equation}
where $p_{ij}=1$ represents the corresponding event for the $i$-{th} input text.  Different from~\cite{DBLP:conf/naacl/LiuLZYZ19} that converts the multi-label classification task into a series of binary classification tasks, our proposal can outputs all event type(s) simultaneously. 


\paragraph{\bf EDM} We will describe the implementation of EDM here. More detailed explanations on the design EDM with supporting experiments are provided in Sec.~\ref{sec:exp_edm}. During training, the event derangement module is proposed to mitigate the imbalanced learning issue. In combinatorics, {\em Derangement} represents a permutation of the elements in a set that makes no elements appear at its original position. In our implementation, one should know that $E_{\tiny\mbox{D}}$ only consists of $r$ events from the beginning of $S_{\tiny\mbox{SA}}$ excluding those in $E_{\tiny\mbox{GT}}$; see line 2 of Algo.~\ref{alg:EDM}.  From line 6 of Algo.~\ref{alg:EDM}, we can know that the derangement procedure is conducted with probability $q$ only when the target (i.e., the ground-truth) events are major events. From line 7-10 of Algo.~\ref{alg:EDM}, we derange the sequence $S_{\tiny\mbox{init}}$ by switching different events in $E_{\tiny\mbox{D}}$. 



{The event derangement module can prohibit the model from excessively learning major events, which works similarly to under-sampling the training instances of major events.  The training process will be more balanced via EDM.}

\section{Experiments}\label{sec:exp}
We present the experimental setups and overall performance in the following.

\begin{table*}[]
\centering
\begin{tabular}{|ll|ccc|ccc|}
\hline
\multirow{2}{*}{\textbf{Methods}} && \multicolumn{3}{|c|}{Subtypes (\%)} & \multicolumn{3}{c|}{Main (\%)} \\\cline{3-8}
                                        &  & \textbf{P}    & \textbf{R}    & \textbf{F1}   & \textbf{P}       & \textbf{R}       & \textbf{F1}      \\ \hline
TBNNAM~\cite{DBLP:conf/naacl/LiuLZYZ19}                     &  & 76.2                           & 64.5                           & 69.9                           & -                & -                & -               \\
TEXT2EVENT~\cite{lu-etal-2021-text2event}     &  & 69.6                             & 74.4                             & 71.9                          & -                & -                & -               \\
DEGREE~\cite{hsu2022degree}              &  & -                           & -                           & 73.3                           & -                & -                & -               \\\hdashline
BERT\_RC\_Trigger~\cite{DBLP:conf/emnlp/DuC20}              &  & 71.7                           & 73.7                           & 72.3                           & -                & -                & -               \\
RCEE\_ER~\cite{DBLP:conf/emnlp/LiuCLBL20}                   &  & 75.6                           & 74.2                           & 74.9                           & -                & -                & -               \\
DMBERT + Boot~\cite{wang2019adversarial}                    &  & 77.9                           & 72.5                           & 75.1                           & -                & -                & -              \\
CLEVE~\cite{DBLP:conf/acl/WangW0L000L020}                           &  & 78.1                           & 81.5                         & 79.8                         & -                & -                & -               \\
 \hline
BERT Finetune                                                &  & 72.8                           & 68.7                           & 70.7                           & 78.0             & 70.8             & 74.2            \\
Our ED\_RC                                                  &  & 76.9                           & 72.3                           & 74.7                           & 78.9             & 75.4             & 77.1            \\
Our ED\_DRC &  & {79.5} & {76.8} & {78.1} & {78.7 }             & { 79.0}             & {78.9 }            \\ \hline
\end{tabular}
\caption{Event detection results on both the event subtypes and event main types of the ACE2005 corpus.} 
\label{tab:evals}
\end{table*}

\subsection{Experimental Setups}\label{sec:setup}

\paragraph{Datasets and Evaluation} 
We conducted experiments on two benchmark datasets:
\begin{compactitem}
\item The {\bf ACE2005} corpus consists of 8 event main types and 33 subtypes~\cite{doddington-etal-2004-automatic}.  {The data distribution of event subtypes is heavily imbalanced (IR$\approx$605.5) as shown in Fig.~\ref{fig:ace_second_level}.  For example, the types of {\em Attack}, {\em Transport}, and {\em Die} account for over half of the total training data.  Meanwhile, the distribution of event main types is more balanced, IR$\approx$13.1 and see Fig.~\ref{fig:ace_first_level}.}  For fair comparison, we follow the evaluation of~\cite{DBLP:conf/acl/LiJH13,DBLP:conf/naacl/LiuLZYZ19,DBLP:conf/emnlp/LiuCLBL20}, i.e., randomly selecting 30 articles from different genres as the validation set, subsequently delivering a blind test on a separate set of 40 ACE2005 newswire documents, and using the remaining 529 articles as the training set. 
\item The {\bf TAC-KBP-2015} corpus~\cite{DBLP:conf/tac/EllisGFKSBS15} is annotated with event nuggets in 38 types.  We process the data following~\cite{DBLP:conf/emnlp/PengSR16}.  The data distribution is more balanced than ACE2005 with IR $\approx 61.5$ as shown in Fig.~\ref{fig:kbp_second_level}.  
\end{compactitem}
The standard metrics: {\bf P}recision, {\bf R}ecall, and {\bf F1} scores, are applied to evaluate the model performance.


\paragraph{Implementation Details}
Our implementation is in PyTorch.  The {\em bert-base-uncased} from Hugging Face~\cite{DBLP:journals/corr/abs-1910-03771} is adopted as the backbone model.  The MLP consists of two layers with the sizes of 768 and the number of event types (e.g., {34 for ACE2005 and 39 for TAC-KBP-2015} including a ``negative'' type) to predict the probability of the input sentence assigned to the corresponding event types.  We follow~\cite{dong2018imbalanced} to set $\alpha$ as 0.5 and round $k$ to the nearest integer based on the calculation by Eq.~(\ref{eq:major_class_definition}).  In EDM, the derangement probability $q$ is set to 0.2.  Based on our validation observation (see more supporting results in Appendix~\ref{sec:exp_q_r}), we set the number of deranged tokens $r$ to 24 for the event subtypes in ACE2005 and TAC-KBP-2015, respectively, and 3 when testing the event main types in ACE2005.  The batch size is set to 8.  The dropout rate is 0.1.  ADAM is the optimizer~\cite{DBLP:journals/corr/KingmaB14} with a learning rate of $2\times 10^{-5}$.  We train our models for 10 epochs to give the best performance.  All experiments are conducted on an NVIDIA A100 GPU in around 1.5 hours.  

\subsection{Overall Performance}
We name our proposal without and with EDM as ED\_RC and ED\_DRC, respectively, and 
compare them with several competitive baselines on ACE2005: in terms of low-resource ED, 1) {\bf TBNNAM}~\cite{DBLP:conf/naacl/LiuLZYZ19}: an LSTM model detecting events without triggers; 2) {\bf TEXT2EVENT}~\cite{lu-etal-2021-text2event}: a sequence-to-structure model that directly learns from parallel text-record annotation that requires no trigger offsets; 3) {\bf DEGREE}~\cite{hsu2022degree}: a generation-based model that leverages manually designed prompts so as to employ only partial data. And for trigger-based ED models, 4) {\bf BERT\_RC\_Trigger}~\cite{DBLP:conf/emnlp/DuC20} and 5) {\bf RCEE\_ER}~\cite{DBLP:conf/emnlp/LiuCLBL20}: both BERT-based models converting event extraction as an MRC task; 6) {\bf DMBERT}~\cite{wang2019adversarial}: a BERT-based model leveraging adversarial training for weakly supervised events, where DMBERT Boot stands for bootstrapped DMBERT; 7){\bf CLEVE}~\cite{DBLP:conf/acl/WangW0L000L020}: a contrastive pre-training framework that exploits the rich event knowledge from large-scale unsupervised data. Besides, the baselines are only evaluated on the event subtypes.

Table~\ref{tab:evals} reports the overall performance on the ACE2005 corpus. It shows that 
\begin{compactitem}
\item Although our proposed ED\_RC does not have access to the triggers, it attains a much better performance than other models for low-resource ED, namely TBNNAM, TEXT2EVENT, and DEGREE.  Its performance is also competitive to DMBERT and RCEE\_ER, reaching 74.7\% F1 score, which is only 0.4\% less than that of the best baseline, DMBERT Boot.  The result shows that our proposed RC framework is capable of learning relations between given texts and event types even without triggers.  
\item After including EDM, our proposed ED\_DRC can significantly outperform all compared methods in all three metrics except CLEVE.  Despite the significant lack of clues, our trigger-free ED\_DRC has a very close F1 score to CLEVE's. Additionally, the F1 score of ED\_DRC gains 3.0\% higher than that of DMBERT Boot. 
\item To verify the consistence of our proposal, we also conduct experiments to evaluate the performance on event main types.  In this case, our proposed ED\_RC and ED\_DRC are both shown to gain greater improvement in F1 score,  which are 2.9\% and 4.7\% better than the finetuned BERT respectively. 
\end{compactitem}

\begin{table}[]
\centering
\begin{tabular}{|l|ccc|}
\hline
\textbf{Methods}       & \textbf{P} & \textbf{R} & \textbf{F1} \\ \hline
\text{BERT Finetune} & 84.1       & 65.0       & 71.7        \\
\text{Our ED\_RC}  & 77.4       & 74.8       & 76.1        \\
\text{Our ED\_DRC} & {79.8}       & {75.2}       & {77.4}        \\ \hline
\end{tabular}
\caption{Event detection results on TAC-KBP-2015.}
\label{tab:kbp}
\end{table}
To further test the generalization of our proposal, the results on a different dataset,TAC-KBP-2015, are reported (see Table~\ref{tab:kbp}) and demonstrate that our ED\_RC can greatly outperform the finetuned base BERT by $4.4\%$ in F1 score.  The event derangement module can further improve our ED\_DRC by $1.3\%$.  But the increment is relatively smaller than that on the ACE2005 dataset.  This can be attributed to more balanced event type distribution on TAC-KBP-2015 than that on ACE2005.  All in all, our ED\_RC is proven capable in both the ACE2005 English dataset and the TAC-KBP-2015 dataset. 

\section{More Analysis}\label{sec:exp_analysis}
{In this section, we will further analyze the effect of each component in our proposal and explain the underlying mechanism.}

\begin{table}[htp]
\centering
\begin{tabular}{|l|ccc|}
\hline
 & \textbf{P} & \textbf{R} & \textbf{F1} \\ \hline
ED\_RC\_Same                       & 75.7      & 71.6      & 73.6       \\ \hdashline
ED\_RC               & 76.9      & 72.3      & 74.7       \\ \hdashline
ED\_RC\_Shuffle\_Test      & 18.2      & 9.2      & 12.2      \\ \hline
ED\_DRC\_Shuffle\_Test      & 66.0      & 45.1      & 53.6       \\ \hdashline
ED\_DRC       & \textbf{79.5} & \textbf{76.8} & \textbf{78.1}      \\ \hline
\end{tabular}
\caption{Evaluation results on different kinds of event orders for ACE2005.}
\label{tab:testSeq}
\end{table}

{
\subsection{Effect of Event Orders}\label{sec:exp_event_sequence} 
Table~\ref{tab:testSeq} explores the effect of the event orders in different cases to understand the underlying mechanism of our proposal.
The first two rows suggest that event orders help our ED\_RC detect events. Those two rows record the results of ED\_RC\_Same and ED\_RC, where ED\_RC\_Same applies the same position embedding to all event types to eliminate the difference in the event orders.  On the contrary, ED\_RC applies varied position embedding to each event type.  The results of the first two rows show that by leveraging the positional difference of event tokens, ED\_RC gains 1.1\% improvement on the F1 score.  


Additionally, our ED\_RC is shown to be event-order-sensitive.  This can be verified by the results of ED\_RC and ED\_RC\_Shuffle\_Test.  Here, ED\_RC\_Shuffle\_Test is trained with the same event order of ED\_RC, but tested with a shuffled event order sequence.  By confusing ED\_RC with a different event order sequence during inference, we obtain a devastating drop on the F1 score, from 74.7\% to 12.2\%.  This implies that our ED\_RC mainly relies on the order of events to recognize them. }

{Furthermore, by performing derangement on ED\_RC\_Shuffle\_Test, we obtain significant performance improvement on ED\_DRC\_Shuffle\_Test.  We conjecture that our proposed event derangement module undermines the excessive reliance of ED\_RC on event orders by introducing perturbation. The model is thus made to learn more about semantics between the input text and the event types, which helps ED\_DRC outperform ED\_RC.
}


\begin{figure}[htp]
\centering
\includegraphics[scale=0.5]{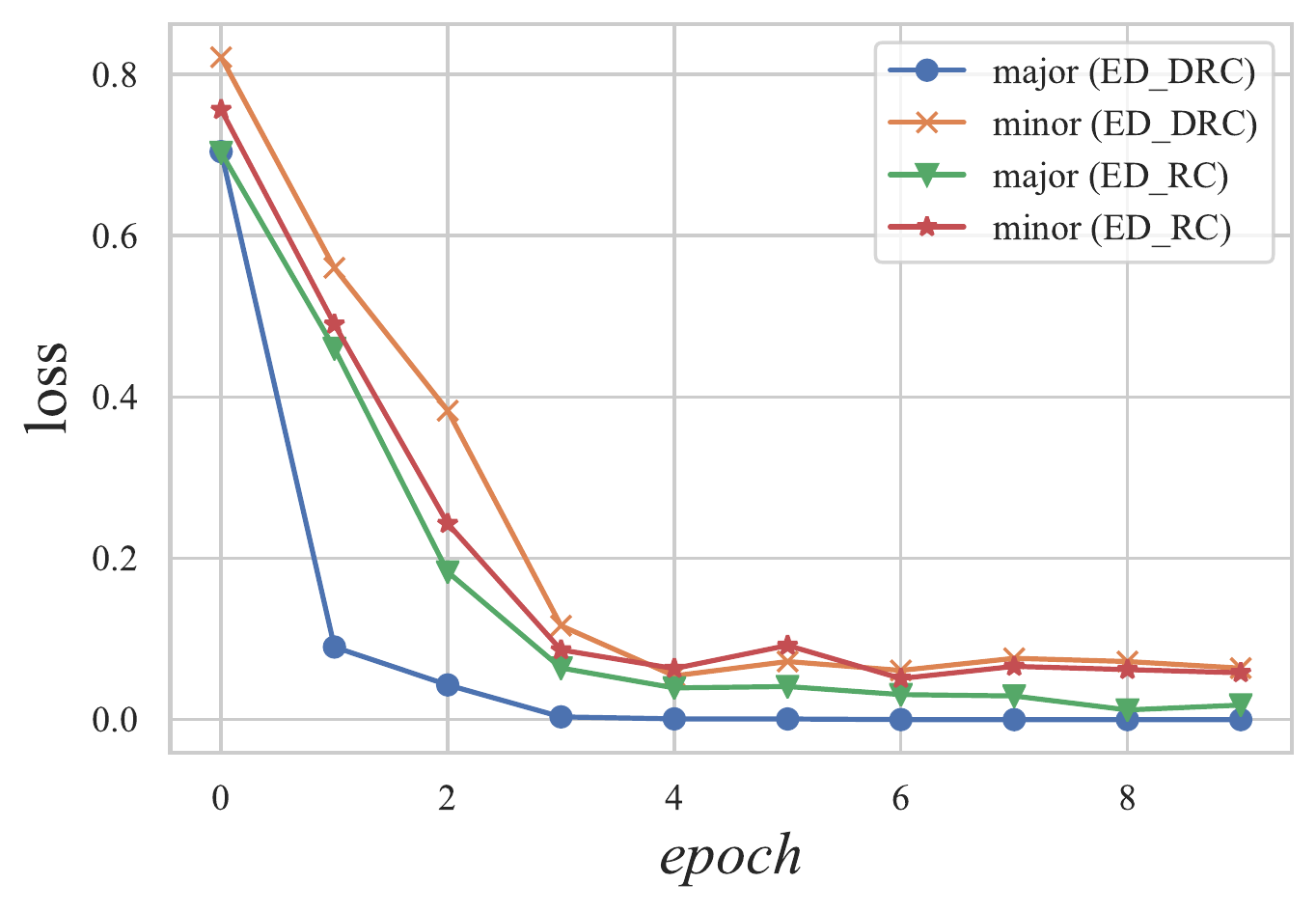}
\caption{{The training losses of ED\_DRC and ED\_RC on ACE2005 with respect to the major events and the minor events, respectively.}  
}
\label{fig:loss}
\end{figure}
\subsection{Effect of EDM}\label{sec:exp_edm}  

Fig.~\ref{fig:loss} shows the training losses of ED\_DRC and ED\_RC on ACE2005 with respect to the major events and the minor events to illustrate how EDM works.  To amplify the effect, we test an extreme case, i.e., setting $q=1.0$, which means that the event derangement will be conducted for each instance whose ground-truth events are major events.  From Fig.~\ref{fig:loss}, we observe that
\begin{compactitem}
\item The solid line with circle markers records the loss of ED\_DRC on the major events and shows that the loss drops much faster and is much smaller (close to zero) than the counterpart of ED\_RC. 
\item The solid line with 'x' markers records the loss of ED\_DRC on the minor events and shows that the loss is relatively higher than the counterpart of ED\_RC before convergence.
\end{compactitem}
 We conjecture that the swift convergence of ED\_DRC on major events is attributed to the positional hint of ground-truth (major) events given by derangement, because our model is sensitive to event orders.  During derangement, the positions of ground-truth (major) events are reserved while some surrounding events are deranged.  Such positional hints help the model to recognize ground-truth events without much effort.  The training loss on the major events is marginal, which yields little gradient update on the major events in ED\_DRC.  The major events are thus prohibited from being excessively learned.  Meanwhile, the relatively higher loss on the minor events makes ED\_DRC focus more on updating the model when the training instances are from the minor events.  In short, our derangement procedure implicitly undersamples the instances of major events during training, which makes the training process more balanced.  {Thanks to EDM, our ED\_DRC increases the F1 score of ED\_RC by 3.4\% as shown in Table~\ref{tab:evals}.  More significantly, the F1 score on the minor events can be improved from 69.1\% to 72.4\%.}

\begin{figure}[htp]
    \centering
    \includegraphics[scale=0.7]{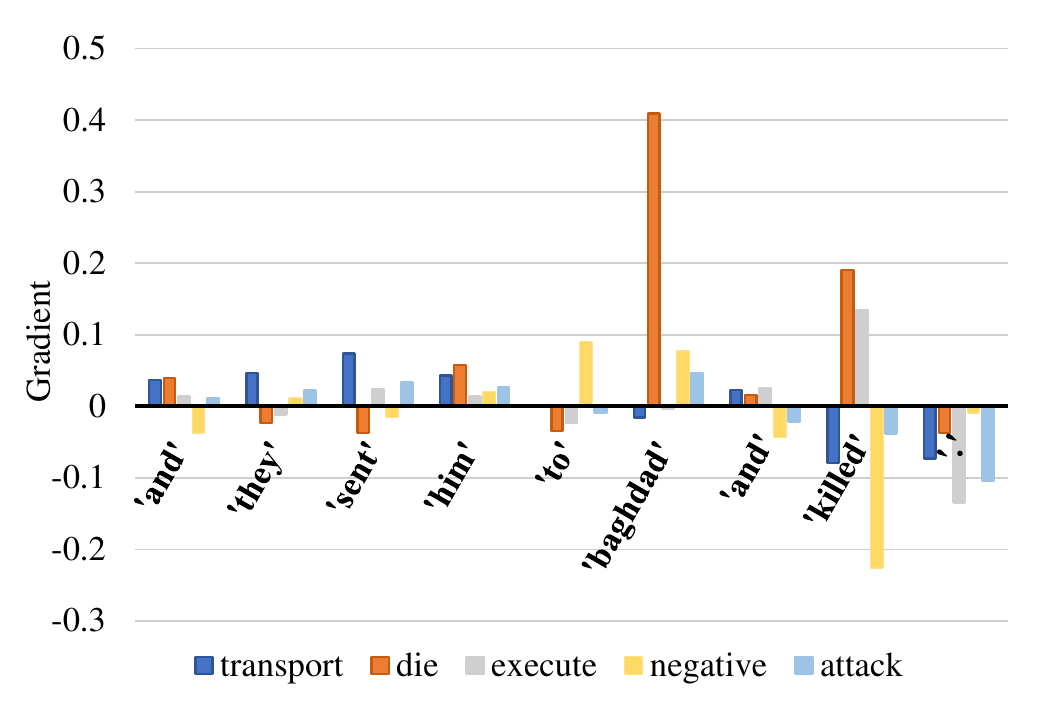}
    \caption{Gradient visualization of words in a sentence from ACE2005 with respect to five typical event types; see more descriptions in the main text.  
    }
    \label{fig:gradient}
\end{figure}

\subsection{Gradient Explanation}\label{sec:exp_ge} 
In order to interpret how our model understands input texts and identifies event types, we computed the gradients with respect to the embeddings of the input text. Those gradients quantify the influence of changes in the tokens on the predictions. In the literature, the gradient explanation has been verified as a more stable method to explain the attention model~\cite{DBLP:conf/nips/AdebayoGMGHK18} than the attention weights in BERT because the attention weights may be misleading~\cite{DBLP:conf/naacl/JainW19} or are not directly interpretable~\cite{DBLP:conf/iclr/BrunnerLPRCW20}.  Here, we pick the example from ACE2005 in Sec.~\ref{sec:intro} and select five events by the following criteria: ``Die'' and ``Transport'' are the target events; ``Negative'' and ``Attack'' are two most common event types; and ``Execute'' is a minor event with semantic closeness to the target events.  Figure~\ref{fig:gradient} clearly shows that 
\begin{compactitem}
\item For the event of ``Die'',  our ED\_DRC can automatically focus on its trigger word ``killed'' while for the event ``Transport'', the trigger ``sent'' is also notified by model.  But for non-target events, our ED\_DRC attains low gradients on the triggers or gets high gradients on unrelated tokens (e.g., ``to'').  
\item Our ED\_DRC can also surprisingly spot the arguments related to the target events.  For example, for the event of ``Die'', ``Baghdad'' yields a significant higher gradient, which corresponds to the argument of PLACE.  Similarly, for the event of ``Transport'', ``they'' and ``him'' also yield relatively larger gradients, which respectively correspond to the argument of ARTIFACT and AGENT.  
\end{compactitem}
{In summary, these observations indicate that our proposed ED\_DRC can successfully link triggers to the corresponding target events and may take related event arguments into consideration as well for identifying events.  More gradient visualization results on the samples from TAC-KBP-2015 are provided in Fig.~\ref{fig:kbp_gradient} of Appendix.~\ref{sec:rs_TAC-KBP} and reveal similar observations as well.}


\section{Conclusion and Future Work}
In this paper, we propose a novel trigger-free event detection method via Derangement mechanism on a machine Reading Comprehension framework.  By adopting BERT in the machine reading comprehension framework, we can absorb the semantic relations between the original input text and the event type(s).  Moreover, the proposed event derangement module can mitigate the imbalanced training process.  Via position perturbation on the major events, we can prevent the major events from being excessively learned while focusing more on updating the model when the training instances are from the minor events.  We conduct empirical evaluation to show that our proposal achieves state-of-the-art performance over previous methods.  Meanwhile, we make further analysis to uncover why our proposal works.  The provided gradient visualization results imply that our proposed trigger-free event detection method can highlight the triggers for the corresponding events by itself and signify the related event arguments as well.


Several promising directions can be considered in the future.  First, since our proposal is event-order-sensitive, it is worthy to explore how to generate an optimal initial event order to further improve the performance.  Second, the gradients of our model with respect to input words have been shown effective to signify triggers and arguments.  It may be meaningful to leverage those gradients to help extract the key information of events, e.g., arguments.  Third, it would be worthwhile to adapt our proposal to other information extraction tasks, e.g., relation extraction, to extend its application scope.  

\bibliography{COLING} 
\bibliographystyle{acl_natbib}

\appendix
\section{Appendix}
\label{sec:appendix}
We provide more analysis to understand our proposal.

\begin{table}[htp]
\centering
\begin{tabular}{|l|ccc|}
\hline
\multicolumn{1}{|l|}{\textbf{Conversion}} & \textbf{P} & \textbf{R} & \textbf{F1} \\ \hline
Original                                    & 75.2       & 67.6       & 71.7        \\
New                              & 77.3       & 68.2       & 72.5        \\ \hline
\end{tabular}
\caption{Results of different conversion ways of event tokens.}
\label{tab:wdembed}
\end{table}
\subsection{Effect of Event Tokens Conversion}
\label{sec:wdembed}

There are two intuitive ways to treat the event tokens in our proposed framework.  One is to treat them as old words in the BERT dictionary, so that we can initialize the event representations by utilizing BERT's pre-trained word embeddings.  The other way is to treat them as new words, so that we can learn the event representations from scratch.  Hence, we can directly feed the {\em original} event words in the DRC framework or add a square bracket around the event words to convert them into {\em new} words, e.g., ``Transport'' to ``[Transport]'', in the BERT dictionary.  

Table~\ref{tab:wdembed} reports the compared results and shows that converting event types into new words can attain substantial improvement in all three metrics than treating them as the original words in BERT dictionary.  We conjecture that it may arise from WordPiece~\cite{DBLP:journals/corr/WuSCLNMKCGMKSJL16} in BERT implementation because BERT will separate an event word into several pieces when it is relatively long.  This brings the difficulty in precisely absorbing the semantic relation between the words in input texts and event types.  On the contrary, when we treat an event word as a new word, BERT will deem them as a whole.  Though BERT learns the event representations from scratch, it is still helpful to establish the semantic relationship between words and event types.

\subsection{Inputs for the Multi-label Classifier}
\label{sec:App_Input_Classifier}

There are two kinds of inputs for the multi-label classifier: the representation of the [CLS] token, or the event representations.  We feed these two inputs into the same MLP to predict the probability of an input sentence $x$ to the corresponding events.
\begin{table}[htp]
\centering
\begin{tabular}{|l|ccc|}
\hline
\textbf{Input}                  & \textbf{P} & \textbf{R} & \textbf{F1} \\ \hline
{[}CLS{]}                          & 77.3       & 68.2       & 72.5        \\ \hline
\multicolumn{1}{|c|}{All event tokens} & 76.9      & 72.3     & 74.7       \\ \hline
\end{tabular}
\caption{Results of different inputs for the multi-label classifier.}
\label{tab:inputCls}
\end{table}

Table~\ref{tab:inputCls} reports the performance of different inputs for the multi-label classifier and shows that by feeding the event representations as the input, our ED\_RC can significantly improve the performance on Recall and the F1 score with competitive Precision score than only using the representation of the [CLS] token.  We conjecture that the event representations have injected more information into the multi-label classifier than only using the representation of the [CLS] token. 

\subsection{Limitation of EDM}\label{sec:appendix-testBalance}
\begin{figure}[htp]
    \centering
    \includegraphics[scale=0.8]{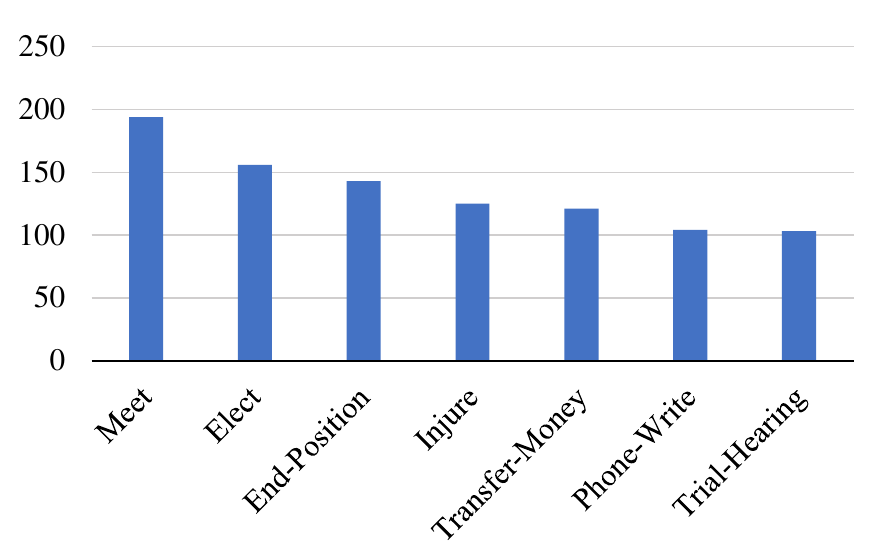}
    \caption{Data distribution of seven balanced event subtypes.}
    \label{fig:balanced}
\end{figure}

We conduct evaluation on a more balanced dataset to investigate the limitation of EDM.  We first select seven relatively balance event types, yielding an imbalance ratio around 1.8, from the subtypes of the ACE2005 corpus; see the data distribution in Fig.~\ref{fig:balanced}.  In the experiment, we set $q$ to 0.2 and $r$ to 6 for good performance on ED\_DRC.

Table~\ref{tab:MajorClsOnly} reports the comparison results of ED\_RC and ED\_DRC and shows that ED\_RC attains satisfactory results and beats ED\_DRC in all three metrics.  The results imply that the derangement procedure plays an important role when the dataset is more imbalanced.  When the dataset is relatively balanced, we can turn to ED\_RC and attain good performance due to the power of self-attention in BERT.

\begin{table}[htp]
\centering
\begin{tabular}{|l|ccc|}
\hline
\textbf{Model} & \textbf{P} & \textbf{R} & \textbf{F1} \\ \hline
ED\_RC  &76.4  &77.8  &77.1 \\
ED\_DRC  &75.0  &76.3  &75.6  \\ \hline
\end{tabular}
\caption{The performance of our ED\_RC and ED\_DRC on a more balanced dataset.}
\label{tab:MajorClsOnly}
\end{table}

\subsection{Error Analysis} \label{sec:app_error_analysis}
We conduct error analysis on test dataset in this section.  There are three main kinds of errors: 
\begin{compactitem}[--]
\item The main error comes from event mis-classification, accounting for $52.9\%$ of the total errors.  The error also includes that ED\_DRC detects more event types than the ground truth.  The most event type that ED\_DRC over-predicts is the event of {\em Attack}.  A typical example is given below:
\begin{quote}
 \textbf{S:} The officials, who spoke on ... \texttt{26 words omitted} ... on the U.S.-backed {\bf war} resolution.
\end{quote}
ED\_DRC deems this sentence belonging to the event of {\em Attack}, where the ground truth is the event of {\em Meet}.  This error is normal because the word ``war'' is a common trigger to the event of {\em Attack}, which yields ED\_DRC mis-classifying it.  In this dataset, the event of {\em Attack} is the most dominating event type, which makes it likely to classify the texts of other events as {\em Attack} when the texts hold some similar words to the triggers of {\em Attack}.  

\item The second type of errors is that ED\_DRC outputs fewer event types than the ground truth, which accounts for $28.9\%$ of errors.  The frequently missing event types are {\em Transfer-Money} and {\em Transfer-Ownership}.  One typical example is 
\begin{quote}
 \textbf{S:} Until Basra, U.S. and British troops ... \texttt{6 words omitted} ... they \textbf{seized} nearby Umm RCsr ... \texttt{3 words omitted} ... \textbf{secure} key oil fields.
\end{quote}
ED\_DRC fails to identify the event of {\em Transfer-Ownership}, which is indicated by the trigger, ``secure'', while recognizing the event of {\em Attack},  implied by the trigger if ``seized''.  On the one hand, the Imbalanced Ratio of {\em Attack} and {\em Transfer-Ownership} is 14.2.  There are much fewer training data for ED\_DRC to learn the patterns of {\em Transfer-Ownership} than those of {\em Attack}.  On the other hand, deeper semantic knowledge is needed for understanding the event of {\em Transfer-Ownership}, whose trigger words are more diverse and changeable.  The triggers for {\em Transfer-Ownership} may include ``sold'',  ``acquire'', and ``bid'', etc.

\item The third type of errors lies in outputting none-event sentences.  When there are no event types in a sentence, ED\_DRC may fail to classify it as the type of {\em negative}.  This is because there is no sufficient clues for ED\_DRC to learn the patterns from the type of {\em negative}.   ED\_DRC also turns out to give low predicted probabilities on all event types.
\end{compactitem}

\begin{figure*}[htp]
    \centering
    \subfigure[ACE2005]
    {\includegraphics[scale=0.5]{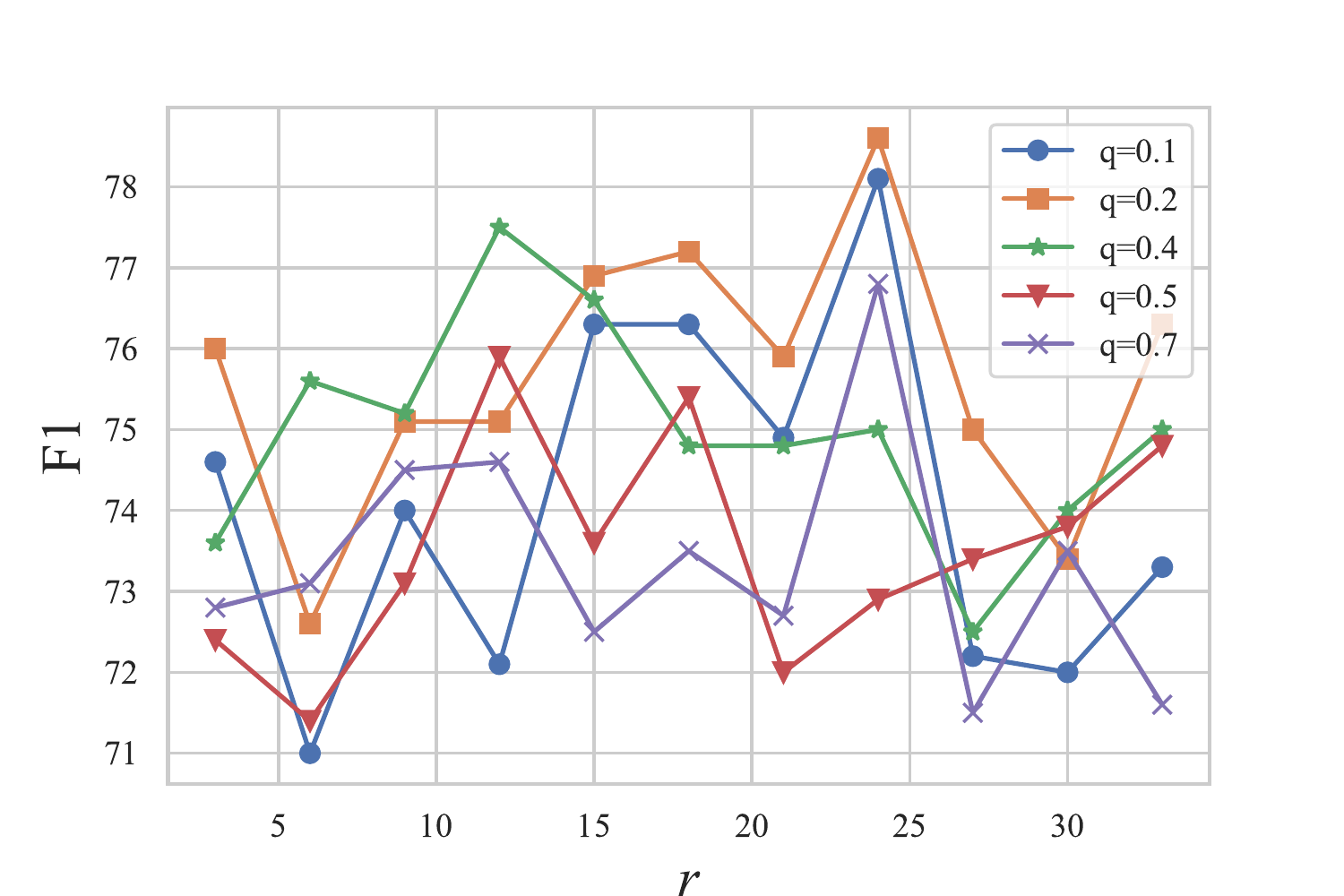}\label{fig:ablation}}
    \subfigure[TAC-KBP-2015]
    {\includegraphics[scale=0.5]{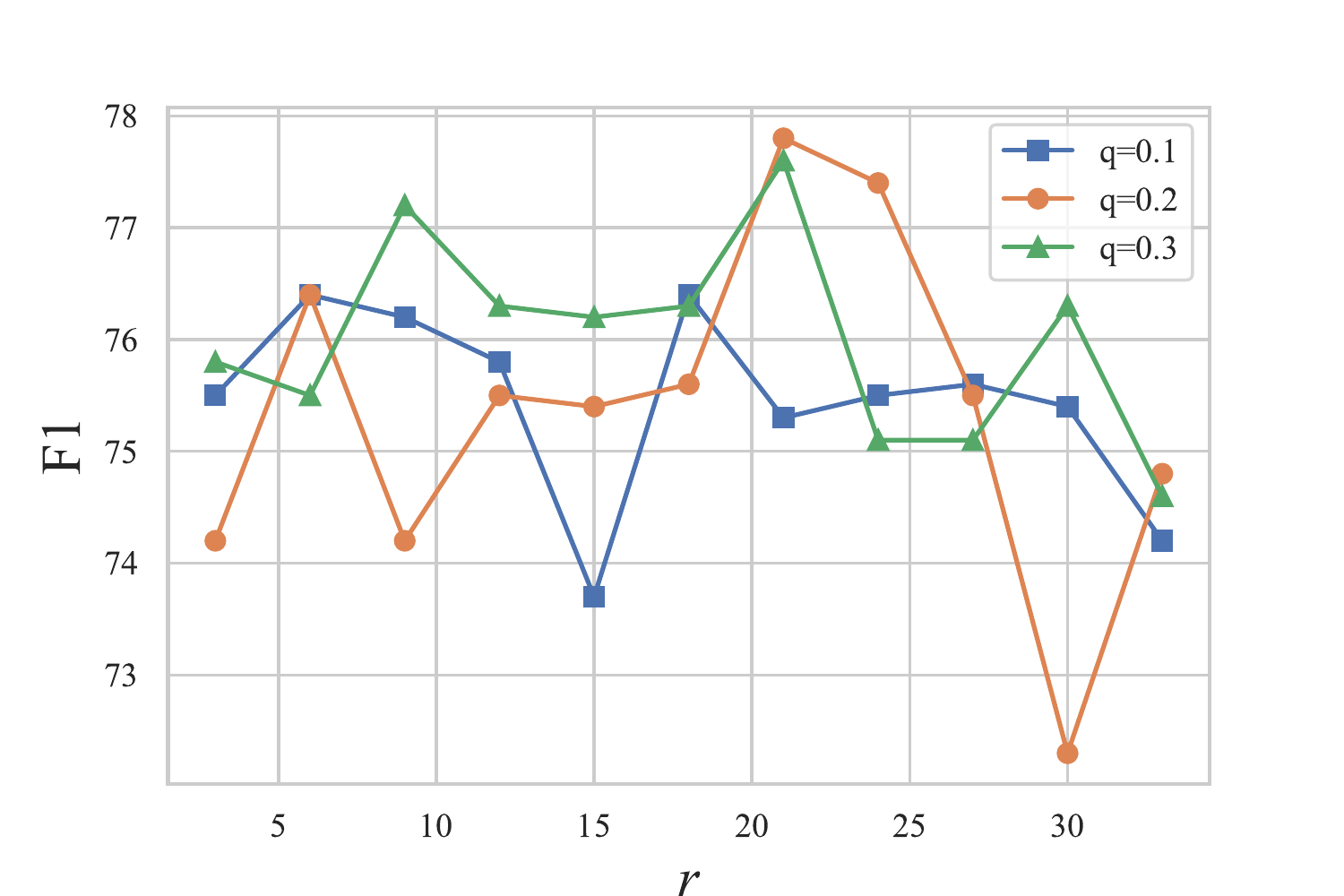}\label{fig:ablation_kbp}}
    \caption{Effect of $q$ and $r$ when evaluating ED\_DRC on ACE2005 and TAC-KBP-2015.}
    \label{fig:ablation_study}
\end{figure*}

\subsection{Effect of Hyperparameters}\label{sec:exp_q_r}  

In this section, we show how to adjust the derangement probability $q$ and the size of the derangement set $r$ by conducting ablation studies where $q$ is selected from $\left\{0.1,\;0.2,\;0.4,\;0.5,\;0.7\right\}$ and $r$ is selected from $\{3, 6, \ldots, 33\}$, i.e., equally dividing all event types into 10 buckets. Larger $q$'s are ignored because they usually fail the model on detecting major events. Figure~\ref{fig:ablation} shows our experiment results on validation set. The best performance is attained when $q=0.2$ and $r=24$ which will then be used during testing.  The trends suggest that a small $q$ should be selected, which can usually yield better performance than a larger one. It is suggested choosing $r$ from 15 to 25, which is approximately from the half to the two-third of the number of total event tokens. A small $r$ may cause negligible perturbation and a relatively large $r$ may affect the disturbance of the minor events since $r$ can determine the scale of perturbation caused by derangement.  

Although these two parameters may be data-oriented, we actually observed similar trends for TAC-KBP-2015. We select $q$ from $\left\{0.1,\;0.2,\;0.3\right\}$ and $r$ from $\{3, 6, \ldots, 33\}$. Figure~\ref{fig:ablation_kbp} shows the performance on TAC-KBP with respect to $r$ for different $q$.  It is shown that the best performance is attained when $q=0.2$ and $r=21$, reaching 77.8\% for F1 score.  The trends remain largely the same as those on the ACE2005 dataset. The best performances also occur when $r$ is selected in the range of 15 and 25. 

\subsection{Gradient Explanation on TAC-KBP-2015}\label{sec:rs_TAC-KBP}

\begin{figure*}[htp]
\centering
\subfigure[``sentence'' as the trigger of the event: ``justice:sentence'' ]{\includegraphics[scale=0.25]{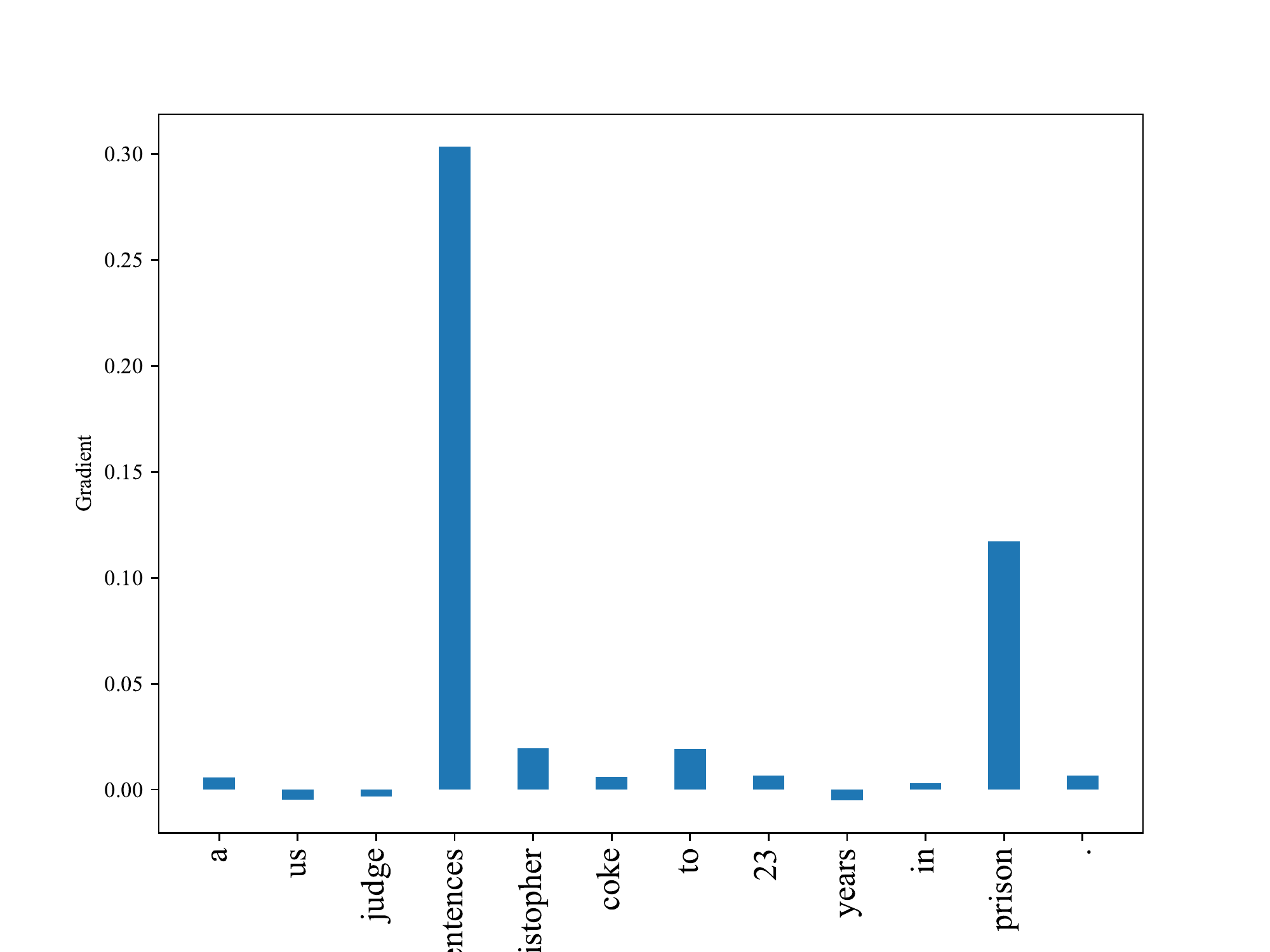}\label{fig:kbp1}}~
\subfigure[``nominated'' as the trigger of the event:  ``transaction:nominate'']{\includegraphics[scale=0.25]{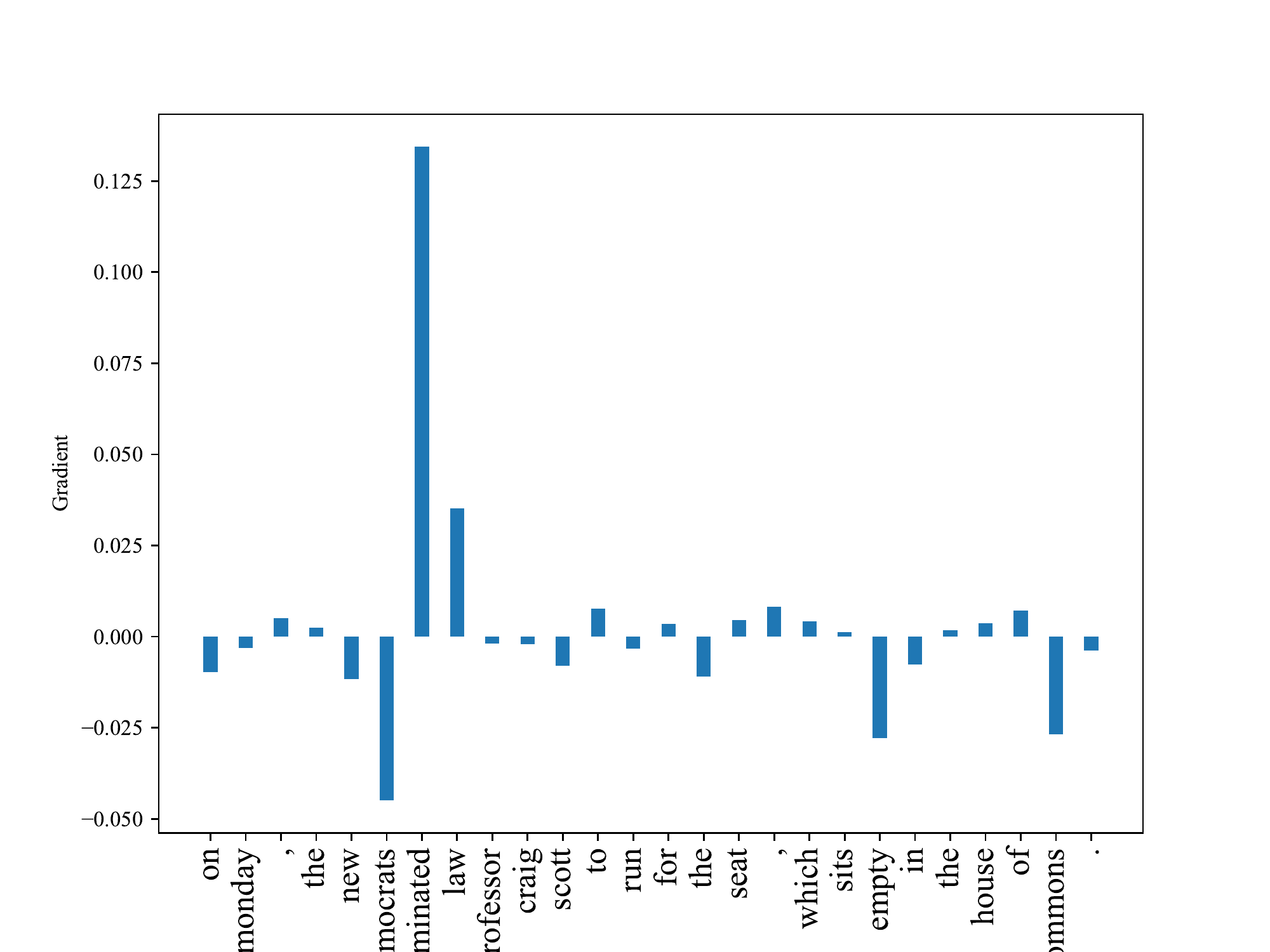}\label{fig:kbp2}}~
\subfigure[``leaked'' as the trigger of the event: ``contact:contact'']{\includegraphics[scale=0.25]{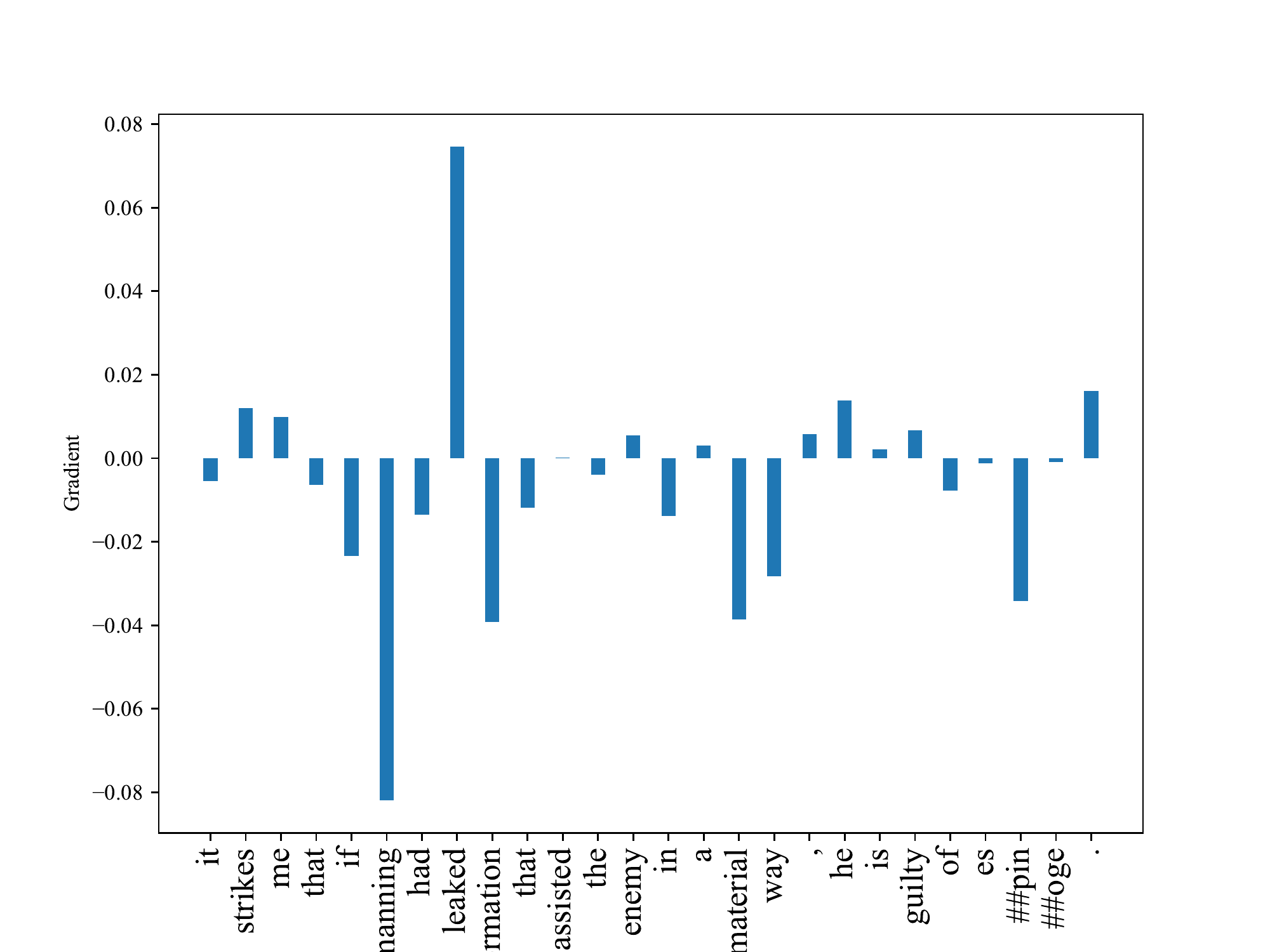}\label{fig:kbp3}}~
\caption{Gradient visualization of words in randomly selected sentences with respect to predicted event types; see more description in the main text.
\label{fig:kbp_gradient}}
\end{figure*}

We conduct gradient explanation on our DRC framework as in Sec.~\ref{sec:exp_ge}.  We randomly choose instances from the test set and visualize gradients respect to the correctly predicted event types by our ED\_DRC.  As shown in Fig.~\ref{fig:kbp_gradient}, ``nominated'', ``leaked'' and ``sentences'' are respectively triggers for those three sentences and receive significant positive gradients compared with other words. This shows that our DRC framework can automatically learn to spot triggers and relate them to event types in practice.

\end{document}